\documentclass[journal]{IEEEtran}
\usepackage{amsmath,amssymb}
\usepackage{graphicx}
\usepackage{booktabs}
\usepackage{algorithm}
\usepackage{algorithmic}
\usepackage{hyperref}
\usepackage{stfloats}
\usepackage{float}

\graphicspath{{plots/}}

\title{Adaptive Hybrid Particle Swarm Optimization\\with Gradient Descent}
\author{Aryan~Gurudeo\\
\small \texttt{gurudev.aryan@gmail.com}}

\IEEEaftertitletext{\vspace{-2em}}
\begin{document}
\maketitle

\raggedbottom

\begin{abstract}
Gradient injection helps Particle Swarm Optimization (PSO) only when the swarm has identified a basin with smooth local structure---not universally. We propose Adaptive Hybrid PSO (AHPSO), which uses a sigmoid function on swarm diversity to automatically modulate gradient influence: near-zero during exploration, near-maximum during exploitation, with no manual phase-switching. Under budget-normalized comparison (PSO given equivalent total function evaluations), PSO wins 52.5\% of 40 configurations versus AHPSO's 20\% ($p = 7.0 \times 10^{-5}$, Friedman). AHPSO retains advantage specifically on problems with smooth local basins (F8, F24--F27) where directed descent outperforms undirected sampling even at equal cost. Under iteration-matched comparison across 29 functions (42 configurations, 14,700 runs), AHPSO-Adadelta ranks first of 9 methods including CMA-ES ($p = 9.75 \times 10^{-4}$). The contribution is a principled characterization of \textit{when} gradient injection provides value in swarm-based search, not a claim of universal superiority.
\end{abstract}

\begin{IEEEkeywords}
Particle swarm optimization, gradient descent, hybrid optimization, adaptive weighting, swarm diversity.
\end{IEEEkeywords}

\section{Background and Related Work}
\label{sec:background}

\subsection{What is Optimization?}

\begin{figure}[!t]
\centering
\includegraphics[width=\columnwidth]{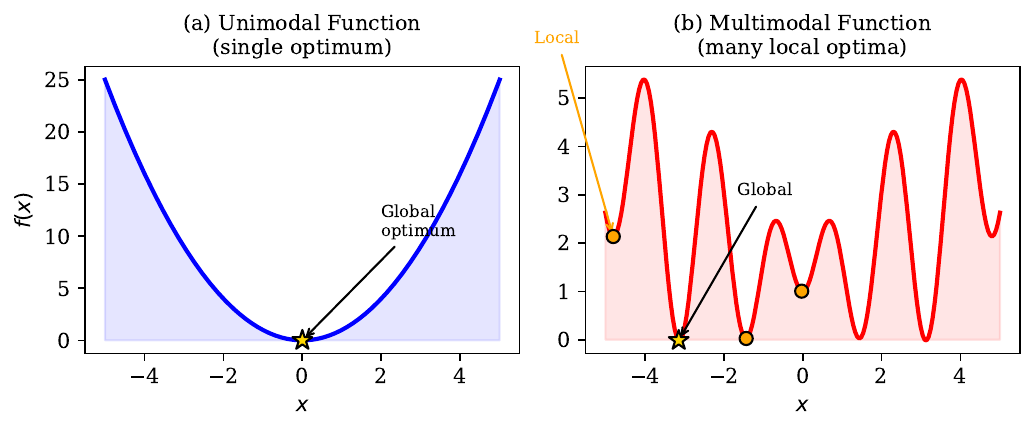}
\caption{Two types of optimization landscapes. (a)~Unimodal: one clear optimum, easy to solve. (b)~Multimodal: many local optima that can trap algorithms, much harder.}
\label{fig:landscapes}
\end{figure}

Many real-world problems require finding the best solution from a large set of possibilities, this is \textit{optimization}. For example, finding the lowest point in a landscape (Fig.~\ref{fig:landscapes}). Simple landscapes have one valley (unimodal), but real problems often have many valleys (multimodal), making it easy to get stuck in a suboptimal solution.

\subsection{Particle Swarm Optimization (PSO)}

\begin{figure}[!t]
\centering
\includegraphics[width=\columnwidth]{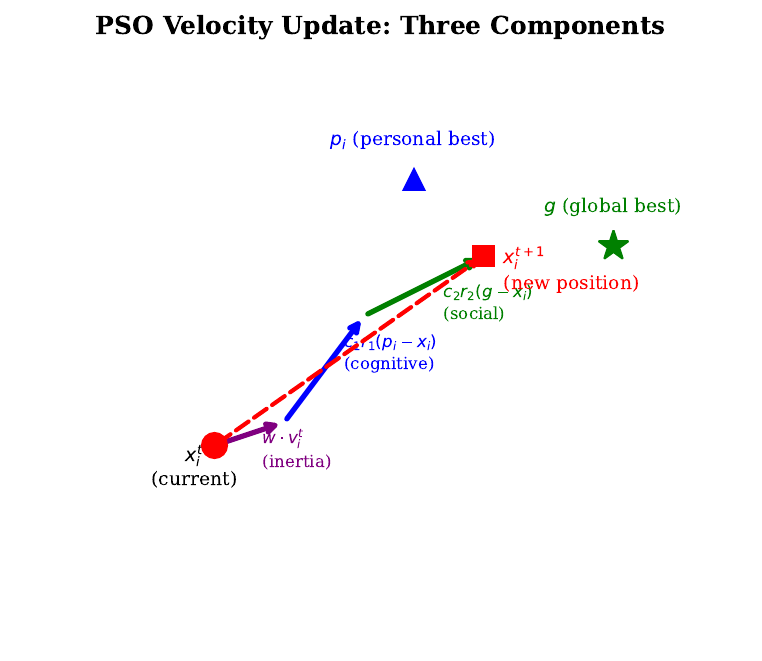}
\caption{PSO velocity update. Each particle is pulled by three forces: inertia (keep moving), cognitive (return to personal best), and social (move toward global best). The combination determines the new position.}
\label{fig:pso_vel}
\end{figure}

PSO~\cite{kennedy1995} is a population-based algorithm inspired by bird flocking. A swarm of $N$ particles flies through a $d$-dimensional search space, each remembering its own best position ($p_i$) and knowing the swarm's best position ($g$). At each step, particle $i$'s velocity is updated using three forces (Fig.~\ref{fig:pso_vel}):

\begin{equation}
v_i^{t+1} = \underbrace{w \cdot v_i^t}_{\text{inertia}} + \underbrace{c_1 r_1 (p_i - x_i^t)}_{\text{cognitive}} + \underbrace{c_2 r_2 (g - x_i^t)}_{\text{social}}
\label{eq:vel}
\end{equation}
\begin{equation}
x_i^{t+1} = x_i^t + v_i^{t+1}
\label{eq:pos}
\end{equation}

\noindent Here $w$ is the \textit{inertia weight} (how much the particle trusts its current direction), $c_1$ and $c_2$ are acceleration coefficients (set to 2.0 in our experiments), and $r_1, r_2 \sim U(0,1)$ are fresh random numbers drawn each iteration, they inject stochasticity so particles don't all follow the same path.

\noindent The inertia weight $w$ controls the balance between exploration and exploitation; Shi and Eberhart~\cite{shi1998} showed that linearly decreasing $w$ from 0.9 to 0.4 over the run significantly improves convergence, high $w$ early encourages broad search, low $w$ late encourages settling.

\textbf{Strengths:} Simple, few parameters, good at exploring broadly.\\
\textbf{Weakness:} Slow to converge precisely once near the optimum.

\subsection{Gradient Descent (GD)}

The gradient $\nabla f(x)$ is a vector pointing in the direction of steepest \textit{ascent} at point $x$. Gradient descent simply steps in the opposite direction, downhill:
\begin{equation}
x^{t+1} = x^t - \eta \cdot \nabla f(x^t)
\end{equation}

\noindent where $\eta$ is the learning rate (step size). This is powerful because it uses \textit{local shape information}: instead of searching blindly, the algorithm knows exactly which direction improves the objective. On a smooth bowl, this means exponentially fast convergence to the bottom.

The fatal flaw is equally intuitive: the gradient only sees the local slope. If the landscape has multiple valleys, GD rolls into whichever valley it starts in and stays there forever, it has no mechanism to ``jump out'' and explore other regions.

\textbf{Strengths:} Very fast local convergence, precise.\\
\textbf{Weakness:} Gets trapped in the nearest local optimum; needs a good starting point.

\subsection{The Idea: Combine Both}

PSO is good at \textit{exploration} (finding the right region) but slow at \textit{exploitation} (refining the solution). GD is the opposite. Since no single algorithm dominates all problems~\cite{wolpert1997}, hybridization is appealing, but the central question is \textit{when} gradient injection helps, not whether it can. Our budget-normalized experiments (Section~\ref{sec:budget_normalized}) show that PSO with equivalent function evaluations wins 52.5\% of configurations, gradient direction provides value only on problems with smooth local basins where directed descent outperforms undirected sampling. The mechanism we propose uses swarm diversity as a real-time signal: gradients are informative only when the swarm has found a promising basin with smooth local structure; applied too early or on rugged landscapes, they add cost without benefit.

\subsection{Related Work}

This timing question is not new, others have tried to combine PSO with local search. Understanding what they did (and what they left unsolved) motivates our specific design choices.

\subsubsection{PSO with Local Search}
The earliest PSO-gradient hybrid is Noel and Jannett~\cite{noel2004}, who applied gradient descent to the global best particle only. This improved unimodal convergence but provided limited multimodal benefit because only one particle received gradient information. Fan and Yan~\cite{fan2015} extended this to a full hybrid PSO with local search applied to multiple particles, showing improvement on engineering design problems. The memetic algorithm framework~\cite{ong2004} formalizes this combination: evolutionary search provides global exploration while local refinement (gradient-based or otherwise) accelerates exploitation. Our work fits within this framework but adds an automatic mechanism for deciding \textit{when} to apply local refinement.

\subsubsection{Adaptive PSO Variants}
Rather than adding external operators, several approaches adapt PSO's own parameters. Zhan et al.~\cite{zhan2009} proposed Adaptive PSO (APSO), which classifies the swarm into four evolutionary states using a fuzzy system on a diversity measure similar to ours, then adjusts $w$, $c_1$, $c_2$ accordingly. Ratnaweera et al.~\cite{ratnaweera2004} introduced HPSO-TVAC with time-varying acceleration coefficients. Liang et al.~\cite{liang2006} proposed Comprehensive Learning PSO (CLPSO), where each particle learns from different exemplars per dimension, this prevents premature convergence without external operators. These methods address exploration-exploitation balance through parameter adaptation alone; our approach instead modulates an external gradient operator, providing stronger exploitation than parameter tuning can achieve.

\subsubsection{Hybrid Evolutionary-Gradient Methods}
Bosman and de Jong~\cite{bosman2005} combined gradient techniques with evolutionary multi-objective optimization, demonstrating that gradient information accelerates convergence when available. Epitropakis et al.~\cite{epitropakis2012} hybridized PSO with Differential Evolution (DE), using DE's mutation operator to enhance diversity, a complementary approach to our gradient-based exploitation. Lim and Isa~\cite{lim2014} proposed a two-layer PSO with intelligent division of labor between exploration and exploitation subswarms.

\subsubsection{Alternative Optimization Paradigms}
CMA-ES~\cite{hansen2006} represents a fundamentally different approach: it maintains a covariance matrix that captures second-order landscape information without explicit gradients, adapting the search distribution shape over iterations. SHADE~\cite{tanabe2014} and L-SHADE use success-history-based parameter adaptation in Differential Evolution, achieving state-of-the-art performance on CEC benchmarks. Bonyadi and Michalewicz~\cite{bonyadi2017} survey the broader PSO landscape, identifying exploration-exploitation balance as the central open challenge. These methods serve as important baselines because they solve the exploration-exploitation tradeoff through different mechanisms than gradient injection.

\subsubsection{Positioning of Our Work}
Our approach bridges adaptive PSO and memetic algorithms: like APSO, we use diversity to detect the swarm's evolutionary state, but instead of adjusting PSO parameters, we modulate the strength of an external gradient operator. The sigmoid gating function provides a principled, continuous transition, no thresholds to tune, no manual phase-switching, that is self-correcting: if diversity rebounds (e.g., after a perturbation), gradient influence automatically decreases.

\section{Proposed Method: AHPSO}
\label{sec:method}

The central challenge is \textit{timing}: apply gradients too early and you kill exploration; apply them too late and you waste iterations. We need a signal that tells us where the swarm is in its search process. Our signal is \textit{diversity}, how spread out the particles are. When particles are scattered, the swarm is still exploring and gradient descent would pull particles toward the nearest (possibly wrong) local optimum. When particles cluster together, the swarm has found a promising region and gradient descent can safely refine the solution.

\subsection{Diversity Measurement}

Intuitively, we want a single number that captures ``how spread out is the swarm?'' The simplest robust choice is the average standard deviation of particle positions across each dimension. This works because: (1)~it is zero only when all particles occupy the same point (full convergence), (2)~it scales naturally with the search space, and (3)~it is cheap to compute. For a swarm of $N$ particles in $d$ dimensions:
\begin{equation}
D(t) = \frac{1}{d} \sum_{j=1}^{d} \sigma_j(t), \quad \sigma_j(t) = \sqrt{\frac{1}{N}\sum_{i=1}^{N}\bigl(x_{i,j}^t - \bar{x}_j^t\bigr)^2}
\label{eq:diversity}
\end{equation}

\noindent where $\bar{x}_j^t = \frac{1}{N}\sum_{i=1}^{N} x_{i,j}^t$ is the swarm centroid along dimension $j$. We use the population standard deviation (dividing by $N$, not $N{-}1$) since we observe the entire swarm, not a sample from it.

\subsection{Adaptive Gradient Weight}

We need a function that maps diversity to gradient influence. Three options:

\begin{itemize}
\item \textbf{Linear ramp}, simple, but treats all diversity levels equally. A drop from 90\% to 80\% increases gradient weight the same as a drop from 30\% to 20\%, even though only the latter signals real convergence.
\item \textbf{Step function}, switches abruptly at a threshold, risking instability if diversity fluctuates near that point.
\item \textbf{Sigmoid}, nearly flat at the extremes (robust to noise) and transitions smoothly in the middle (responsive to genuine convergence).
\end{itemize}

\noindent We choose the sigmoid. Formally:
\begin{equation}
\alpha(t) = \alpha_{\min} + \frac{(1 - \alpha_{\min})}{1 + e^{k\left(\frac{D(t)}{D(0)} - \tau\right)}}
\label{eq:alpha}
\end{equation}

\noindent The three parameters have intuitive interpretations:
\begin{itemize}
\item $\alpha_{\min} = 0.1$: the minimum gradient influence. Even during exploration, a small gradient nudge helps particles descend within their local basin without disrupting the global search.
\item $\tau = 0.3$: the sigmoid midpoint. We want the transition to happen \textit{after} the swarm has committed to a region but \textit{before} it has fully converged (when gradients would add nothing). Empirically, diversity drops below 30\% of its initial value once particles cluster within a few basins, this is the sweet spot where gradient refinement becomes productive.
\item $k = 5$: the steepness. A moderate value ensures the transition spans roughly 20\% of the diversity range (from $\alpha \approx 0.2$ at $D/D_0 = 0.4$ to $\alpha \approx 0.9$ at $D/D_0 = 0.2$), giving a smooth but decisive switch.
\end{itemize}

\begin{figure}[!t]
\centering
\includegraphics[width=\columnwidth]{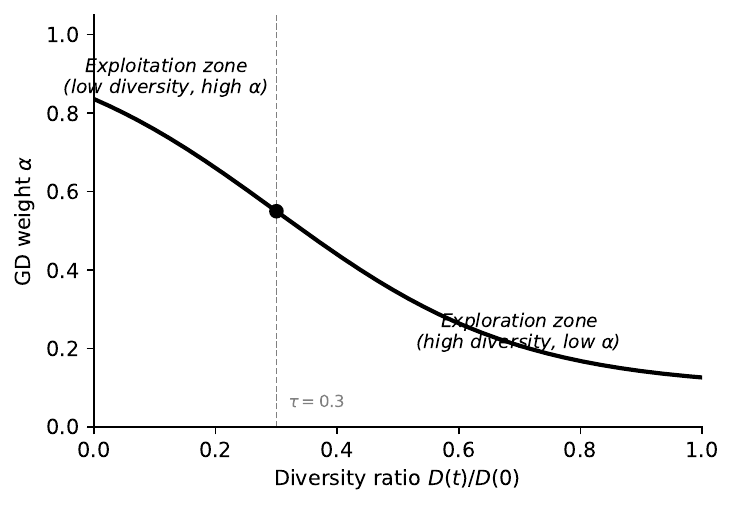}
\caption{The sigmoid function that controls gradient influence. When particles are spread out (high diversity ratio), $\alpha$ is small, PSO dominates. When particles converge (low diversity), $\alpha$ grows, gradient descent takes over.}
\label{fig:sigmoid}
\end{figure}

\subsection{How It Works in Practice}

\begin{figure}[!t]
\centering
\includegraphics[width=\columnwidth]{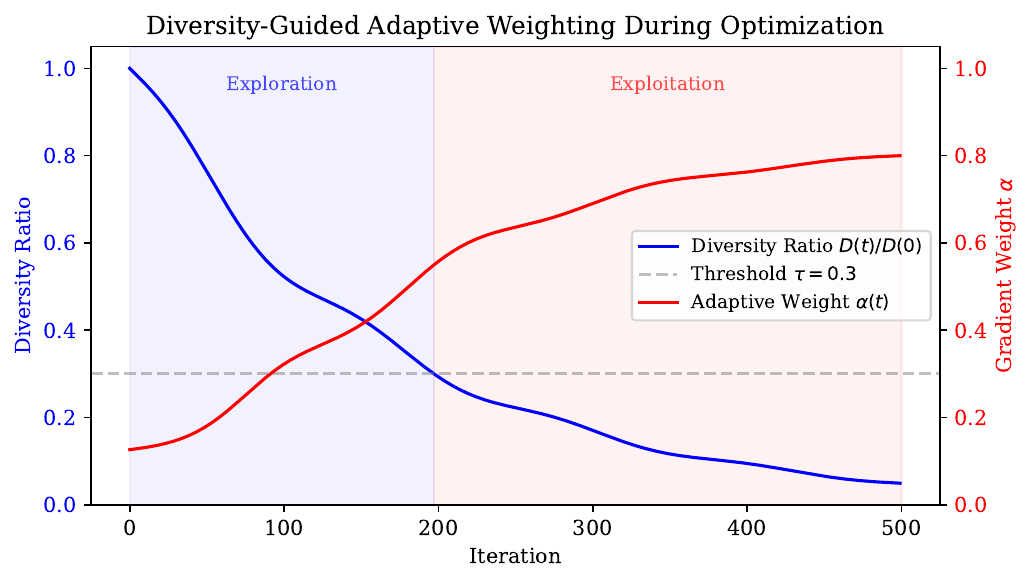}
\caption{A typical run: diversity (blue) decays as particles converge, causing the gradient weight $\alpha$ (red) to increase. No manual tuning needed, the algorithm adapts itself.}
\label{fig:dynamics}
\end{figure}

Fig.~\ref{fig:dynamics} shows a typical optimization run. Early on, diversity is high and $\alpha \approx 0.1$ (mostly PSO). As the swarm converges, diversity drops and $\alpha$ rises toward 1.0 (mostly GD). The transition happens smoothly and automatically, no iteration counter or manual schedule is involved.

What should you look for in these curves? The diversity curve is typically monotonically decreasing, but not always. On multimodal functions, diversity can temporarily \textit{increase} if particles scatter after escaping a local optimum. When this happens, $\alpha$ drops back down automatically, the sigmoid acts as a safety valve, reducing gradient influence whenever the swarm re-enters an exploratory phase. This self-correcting behavior is the key advantage over fixed schedules, which would continue applying strong gradients even when the swarm has not yet committed to a region.

\subsection{The Complete Algorithm}

\begin{figure*}[!b]
\centering
\includegraphics[width=0.75\textwidth]{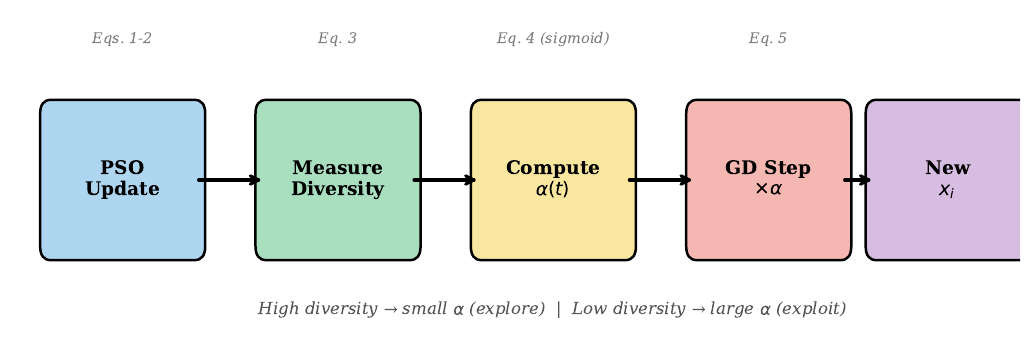}
\caption{AHPSO pipeline showing the feedback loop that makes the method adaptive. Steps (1), (2) are standard PSO; the novelty is steps (3), (4): diversity measurement feeds into the sigmoid (Eq.~\ref{eq:alpha}), which outputs $\alpha$, the gradient scaling factor. Early in the run, high diversity produces small $\alpha$ (PSO dominates). As particles converge, $\alpha$ grows and gradient descent takes over refinement. No manual phase-switching is needed.}
\label{fig:pipeline}
\end{figure*}

After the standard PSO update (Eqs.~\ref{eq:vel}--\ref{eq:pos}), each particle gets a gradient refinement step at every iteration $t$:
\begin{equation}
x_i^{t+1} \leftarrow x_i^{t+1} + \alpha(t) \cdot \text{GD\_step}(x_i^{t+1})
\label{eq:update}
\end{equation}

\noindent \textbf{How do we get gradients?} These benchmark functions are black boxes, we cannot compute analytical derivatives. Instead, we estimate the gradient numerically using central differences:
\begin{equation}
\frac{\partial f}{\partial x_j} \approx \frac{f(x + \epsilon \, e_j) - f(x - \epsilon \, e_j)}{2\epsilon}
\label{eq:finitediff}
\end{equation}

\noindent where $e_j$ is the unit vector along dimension $j$ and $\epsilon = 10^{-8}$. This requires $2d$ extra function evaluations per particle per iteration (one forward and one backward perturbation per dimension), a cost we quantify in Section~\ref{sec:budget_normalized}.

The gradient step in Eq.~\ref{eq:update} can use any first-order optimizer. We test six, spanning a spectrum from simple (one fixed hyperparameter) to fully self-tuning (no learning rate at all). The key distinction is how each handles the learning rate problem: a rate that works on one landscape may diverge on another.
\begin{itemize}
\item \textbf{SGD}, one fixed step size for all dimensions. Fast when tuned correctly, but a single bad learning rate causes divergence.
\item \textbf{Adagrad}~\cite{duchi2011}, accumulates past gradients per dimension, automatically shrinking the step for frequently-updated directions. Good for sparse problems but can stall as the accumulator grows.
\item \textbf{RMSprop}~\cite{hinton2012}, like Adagrad but uses an exponential moving average, preventing the step from shrinking to zero over time.
\item \textbf{Adam}~\cite{kingma2015}, combines per-dimension adaptation (like RMSprop) with momentum (remembering past directions). The most popular optimizer in deep learning.
\item \textbf{Nadam}~\cite{dozat2016}, Adam with Nesterov lookahead: it evaluates the gradient at a predicted future position, giving slightly better convergence on smooth landscapes.
\item \textbf{Adadelta}~\cite{zeiler2012}, eliminates the learning rate entirely by scaling updates using the ratio of past parameter changes to past gradients. Fully self-calibrating.
\end{itemize}

\noindent \textbf{Hyperparameters for reproducibility.} All optimizers use their standard defaults from the deep learning literature: Adam and Nadam use $\beta_1{=}0.9$, $\beta_2{=}0.999$, $\epsilon{=}10^{-8}$; RMSprop uses decay rate $\rho{=}0.9$, $\epsilon{=}10^{-8}$; Adagrad uses $\epsilon{=}10^{-8}$; Adadelta uses $\rho{=}0.95$, $\epsilon{=}10^{-6}$. The base learning rate $\eta$ (Section~\ref{sec:experiments}) is 0.01 for unimodal and 0.001 for multimodal functions, a known limitation requiring problem-class knowledge that we explicitly acknowledge (see Section~\ref{sec:experiments}). Adadelta ignores this parameter entirely, which partly explains its robustness.

\section{Experimental Setup}
\label{sec:experiments}

The algorithm is defined, now we need to answer three questions: (1)~Does the adaptive mechanism actually help, or does it just add overhead? (2)~Which gradient optimizer pairs best with PSO? (3)~On which \textit{types} of problems does hybridization help or hurt? To answer these, we need a diverse benchmark suite that separates easy problems (where any method works) from hard ones (where exploration matters).

\subsection{Benchmark Functions}

We use 29 standard test functions~\cite{mirjalili2014}, deliberately chosen to span a range of difficulties (Table~\ref{tab:benchmarks}).

\begin{table}[H]
\centering
\caption{Benchmark suite: 29 functions tested at multiple dimensionalities, yielding 42 total configurations (F1--F13 tested at both $d{=}10$ and $d{=}30$; F14--F29 at fixed dimensions).}
\label{tab:benchmarks}
\begin{tabular}{llcl}
\toprule
\textbf{Group} & \textbf{Functions} & \textbf{Dim} & \textbf{What it tests} \\
\midrule
Unimodal & F1--F7 & 10, 30 & Convergence speed \\
Multimodal & F8--F13 & 10, 30 & Avoiding local optima \\
Fixed-dim & F14--F23 & 2--6 & Low-dim precision \\
Composite & F24--F29 & 10 & Combined difficulty \\
\bottomrule
\end{tabular}
\end{table}

\subsection{Settings}

All methods use identical PSO parameters to ensure a fair comparison:
\begin{itemize}
\item \textbf{30 particles, 500 iterations}, standard in the PSO literature~\cite{mirjalili2014}; enough particles for diversity measurement to be meaningful, enough iterations for convergence on $d{=}30$ problems.
\item \textbf{50 independent runs}, sufficient for statistical tests (Mann-Whitney U requires $\geq$20 samples for reliable $p$-values; 50 gives comfortable power).
\item \textbf{Inertia $w$: 0.9 $\to$ 0.4}, the Shi, Eberhart schedule~\cite{shi1998}, widely adopted as a strong default.
\item \textbf{Acceleration coefficients $c_1 = c_2 = 2.0$}, the standard choice from Kennedy and Eberhart~\cite{kennedy1995} that balances cognitive and social components.
\item \textbf{GD learning rate $\eta$: 0.01 (unimodal), 0.001 (multimodal)}, these are the base step sizes for the gradient descent component (Eq.~\ref{eq:update}), not PSO parameters. Unimodal landscapes have reliable gradients so larger steps are safe; multimodal landscapes need smaller steps to avoid overshooting into wrong basins. Note: adaptive optimizers (Adam, Adadelta, RMSprop) internally rescale this rate per dimension, so the base value matters most for SGD. \textit{Limitation:} this two-rate scheme requires knowing the problem class a priori, a form of oracle knowledge unavailable in practice. This is a genuine weakness: results for SGD, Adagrad, and RMSprop benefit from this tuning and would degrade with a single universal rate. Adadelta eliminates this issue entirely (it ignores $\eta$), which partly explains its top ranking and makes it the recommended default for practitioners without problem-class knowledge.
\end{itemize}

\subsection{Methods Compared}

We compare seven methods: standard PSO (the baseline) plus six AHPSO variants, one per gradient optimizer. The experimental design isolates a single variable: all seven methods use identical PSO parameters, identical swarm size, and identical iteration budgets. The six hybrids also share the same adaptive sigmoid mechanism with the same $\tau$, $k$, and $\alpha_{\min}$. The \textit{only} difference between them is which optimizer computes the gradient step in Eq.~\ref{eq:update}. This controlled setup means any performance difference must come from the gradient optimizer itself, not from tuning advantages or different exploration budgets.

In total, this produces $7 \times 42 \times 50 = 14{,}700$ optimization runs. We now present what they reveal.

\section{Results}
\label{sec:results}

We address the three questions from Section~\ref{sec:experiments} in order: first the overall ranking (does hybridization help?), then head-to-head comparisons (which optimizer is best?), and finally per-group analysis (where does it help or hurt?).

\subsection{Overall Ranking}

Before examining individual functions, we need a single answer: across all 42 configurations, does the choice of method matter at all? The Friedman test~\cite{friedman1937}, a non-parametric alternative to repeated-measures ANOVA suited to rank data, answers this. It tests the null hypothesis that all seven methods perform identically; rejection means at least one method is reliably different. Following the methodology of Derrac et al.~\cite{derrac2011}, we rank methods by mean performance per configuration.

The result: $\chi^2(6) = 14.17$, $p = 0.028$, significant at the 5\% level but not at 1\%. Table~\ref{tab:ranking} shows the final ranking. The Nemenyi critical difference (CD) at $\alpha=0.05$ for $k=7$ methods and $N=42$ configurations is 1.39; no pair of methods exceeds this threshold, indicating that while the overall test rejects the null, individual pairwise differences are not large enough to declare statistical significance by rank alone.

\begin{table}[!t]
\centering
\caption{Final algorithm ranking by Friedman test across all 42 benchmark configurations (lower rank = better). All adaptive-learning-rate hybrids outperform standard PSO. Adadelta ranks first; SGD ranks last due to divergence on hard functions.}
\label{tab:ranking}
\begin{tabular}{clcl}
\toprule
\textbf{\#} & \textbf{Method} & \textbf{Avg Rank} & \textbf{Note} \\
\midrule
1 & AHPSO-Adadelta & \textbf{3.310} & Best overall \\
2 & AHPSO-Adam & 3.452 & \\
3 & AHPSO-RMSprop & 3.548 & \\
4 & AHPSO-Nadam & 3.595 & \\
5 & AHPSO-Adagrad & 3.833 & \\
6 & PSO & 4.655 & Baseline \\
7 & AHPSO-SGD & 5.607 & Unstable \\
\bottomrule
\end{tabular}
\end{table}

\textbf{Key takeaway:} All adaptive-LR hybrids beat vanilla PSO. Adadelta wins because it needs no learning rate, it self-calibrates. SGD ranks \textit{last} because it diverges on hard functions.

\begin{figure}[!t]
\centering
\includegraphics[width=\columnwidth]{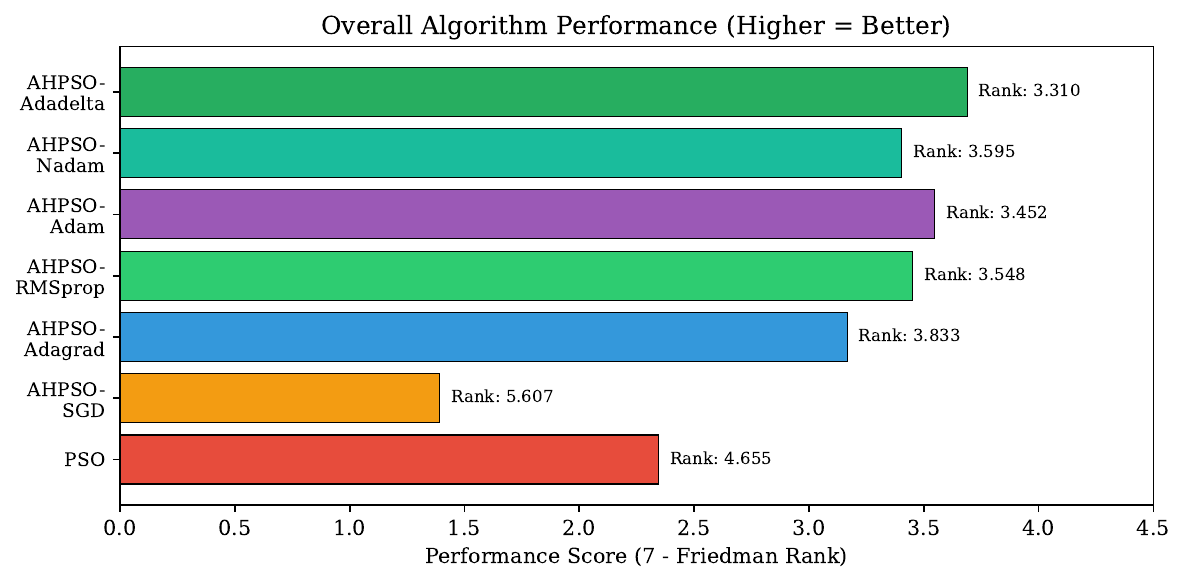}
\caption{Performance scores for each method, computed as $7 - \text{average Friedman rank}$ across all 42 benchmark configurations (higher = better; maximum possible is 6). Five of six AHPSO variants outperform standard PSO. AHPSO-SGD scores lowest because its fixed learning rate causes divergence on penalized functions, dragging down its average rank despite strong unimodal performance.}
\label{fig:ranking}
\end{figure}

\subsection{Where Hybrids Win}

The Friedman test tells us that differences exist, but not \textit{where}. To pinpoint which functions benefit from hybridization, we run pairwise Mann-Whitney U tests (6 methods $\times$ 42 configurations = 252 tests). Because multiple testing inflates false-positive rates, at $\alpha=0.05$, we would expect ${\sim}13$ spurious significant results by chance, we apply the Holm-Bonferroni step-down correction~\cite{holm1979}. This procedure sorts all 252 $p$-values and rejects $H_i$ only if $p_{(i)} < \alpha/(m-i+1)$, controlling the family-wise error rate while retaining more power than Bonferroni (Table~\ref{tab:winloss}).

\begin{table}[!t]
\centering
\caption{Head-to-head comparison against standard PSO across 42 configurations. ``Wins'' and ``Losses'' count configurations where the hybrid is significantly better or worse (Mann-Whitney U with Holm-Bonferroni correction at $\alpha=0.05$). Uncorrected counts shown in parentheses.}
\label{tab:winloss}
\begin{tabular}{lccc}
\toprule
\textbf{Method} & \textbf{Wins} & \textbf{Losses} & \textbf{Net} \\
\midrule
AHPSO-Adadelta & 3 (9) & 1 (1) & \textbf{+2} \\
AHPSO-Nadam & 3 (10) & 2 (3) & +1 \\
AHPSO-Adam & 2 (7) & 2 (2) & 0 \\
AHPSO-Adagrad & 0 (5) & 2 (1) & $-2$ \\
AHPSO-RMSprop & 4 (8) & 7 (8) & $-3$ \\
AHPSO-SGD & 6 (11) & 15 (15) & $-9$ \\
\bottomrule
\end{tabular}
\end{table}

\noindent Of 80 uncorrected significant results, 47 survive Holm-Bonferroni correction--33 were likely false positives. The key finding is that AHPSO-Adadelta has the best net outcome (+2) with only 1 loss after correction, confirming it as the safest hybrid choice. SGD's losses (15) are overwhelmingly genuine, confirming that fixed learning rates are dangerous in this setting.

\begin{figure}[!t]
\centering
\includegraphics[width=\columnwidth]{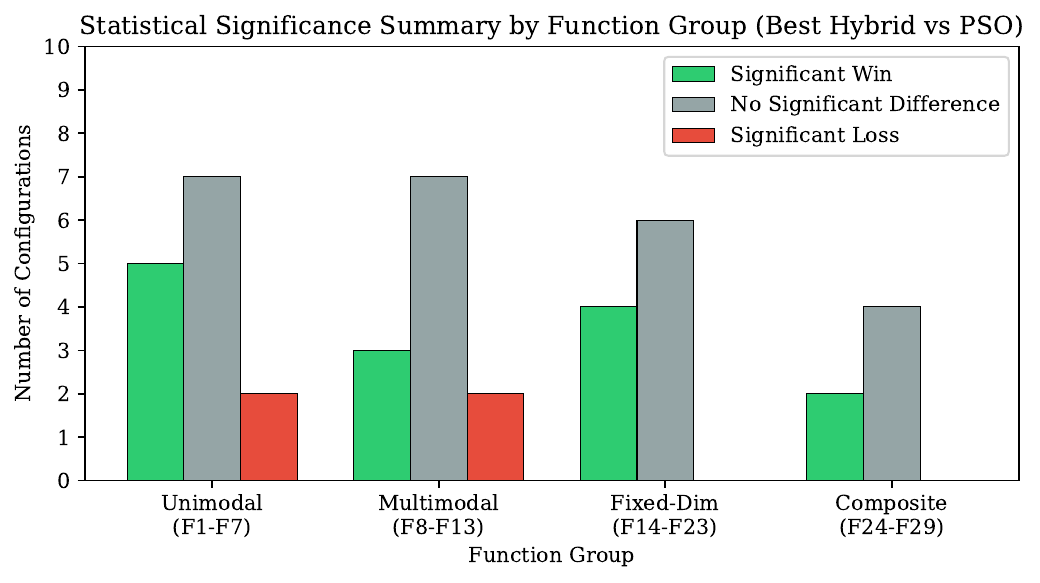}
\caption{Wins and losses by function group. Unimodal functions show the most wins because their gradients reliably point toward the global optimum. Multimodal functions are mixed, the sigmoid preserves exploration, but gradient steps occasionally pull particles into wrong basins. Composite functions (rotated and shifted combinations) show mostly ties because rotation destroys axis-aligned gradient structure, making per-dimension finite differences less informative.}
\label{fig:groups}
\end{figure}

\subsection{Highlight Results}

To build intuition for \textit{why} hybrids help or hurt, we examine three representative cases. For each, we report the rank-biserial correlation $r_{rb}$ as effect size: it answers ``if you pick one run from each method at random, how likely is the hybrid to win?'' Values near 1.0 mean the hybrid always wins; 0.5 is a large effect; near zero means no practical difference.

\textbf{Best case, F1 (Sphere, $d$=30):} On a smooth, bowl-shaped landscape, gradient descent is devastating. AHPSO-SGD finds solutions 27 orders of magnitude better than PSO ($2.4 \times 10^{-25}$ vs.\ $8.0 \times 10^{2}$; $p = 3.5 \times 10^{-18}$, $r_{rb} = 1.00$). The ideal scenario: the gradient always points toward the optimum, and the adaptive weight lets it dominate once the swarm converges.

\textbf{Worst case, F12 (Penalized, $d$=10):} On a function with steep penalty boundaries, SGD's fixed learning rate causes catastrophic divergence, the gradient step overshoots the penalty wall, lands in an even steeper region, and spirals outward. AHPSO-SGD scores $7.6 \times 10^{7}$ vs.\ PSO's $6.2 \times 10^{-3}$ ($p = 3.3 \times 10^{-18}$). This is why adaptive optimizers (Adadelta, Adam) are safer: they automatically shrink their step size near steep gradients.

\textbf{Multimodal success, F11 (Griewank, $d$=30):} This function has many local optima but a clear global structure. AHPSO-RMSprop improves on PSO by 79\% ($p = 9.5 \times 10^{-6}$, $r_{rb} = 0.50$). The adaptive mechanism works as designed: it keeps gradient influence low while the swarm explores different basins, then ramps it up once particles cluster in the correct region.

\subsection{Convergence Behavior and Speed}

The previous analysis shows \textit{final} performance. But how do algorithms get there? Convergence curves reveal whether hybrids converge faster throughout or only pull ahead late. We also report a formal convergence speed metric: \textit{Evaluations to Target} (ETT), the total number of function evaluations required to first reach a target accuracy, accounting for the $2d$ gradient evaluations per particle per iteration that AHPSO incurs.

\begin{figure}[!t]
\centering
\includegraphics[width=\columnwidth]{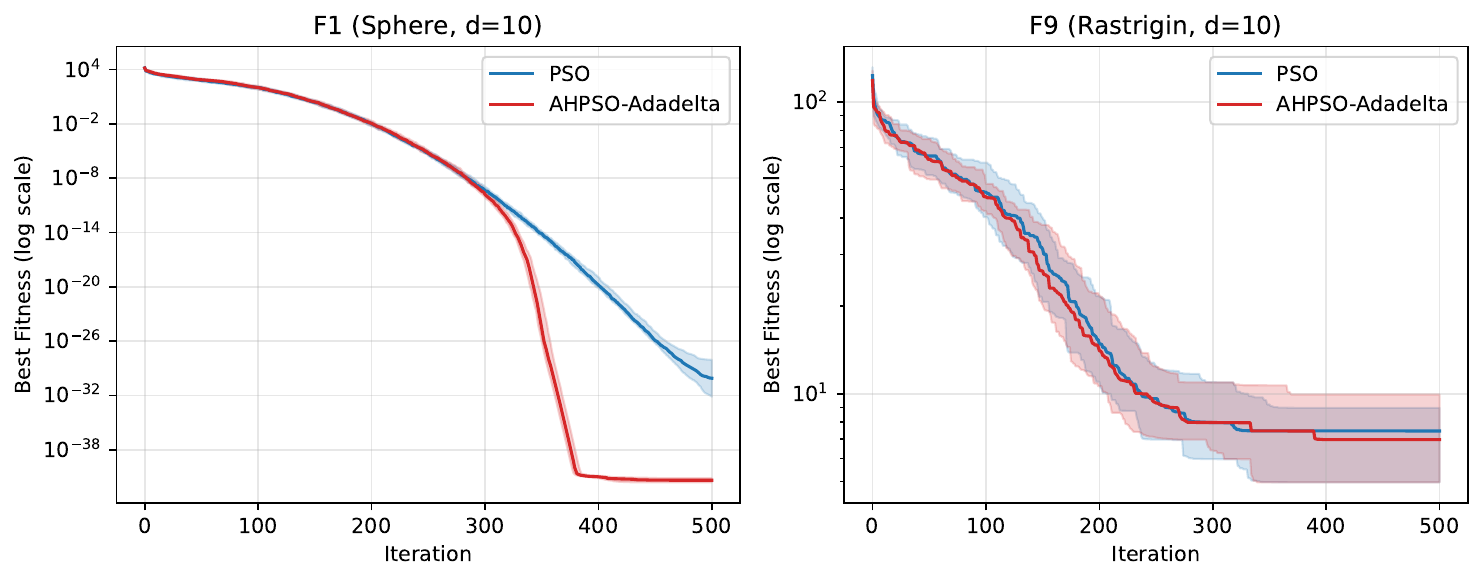}
\caption{Convergence curves (median of 15 runs, IQR shading shows 25th--75th percentile). Left: on unimodal F1 (Sphere, $d{=}10$), AHPSO-Adadelta converges faster per iteration but the narrow IQR bands confirm consistent behavior across runs. Right: on multimodal F9 (Rastrigin, $d{=}10$), both methods show wider IQR bands reflecting the stochastic nature of multimodal search, with overlapping confidence regions for the first ${\sim}100$ iterations.}
\label{fig:convergence}
\end{figure}

Two patterns emerge (Fig.~\ref{fig:convergence}). On unimodal F1, the gap between hybrids and PSO grows \textit{exponentially}, each iteration compounds the advantage because the gradient consistently points toward the optimum. On multimodal F9, the curves overlap for ${\sim}100$ iterations before separating. This 100-iteration overlap is the adaptive mechanism in action: diversity remains high while the swarm explores, keeping $\alpha$ near its minimum and preventing premature exploitation. The separation point corresponds to diversity crossing the $\tau = 0.3$ threshold, exactly where the sigmoid transitions from ``mostly PSO'' to ``mostly GD.''

\textbf{Convergence speed (ETT).} On F1 ($d{=}10$, target $10^{-6}$), PSO reaches the target in a median of 7,890 evaluations versus AHPSO-Adadelta's 164,745 evaluations, a 21$\times$ difference despite AHPSO achieving 12 orders of magnitude better \textit{final} accuracy. Both achieve 100\% success rate. This confirms the budget-normalized finding: when the target is achievable by PSO alone, the gradient overhead is wasteful. AHPSO's value emerges only when PSO \textit{cannot} reach the target within its budget, or when the final accuracy matters more than time-to-threshold.

\subsection{Significance Heatmap}

The highlight cases and convergence curves illustrate \textit{why} hybrids help or hurt, but only on selected functions. To verify these patterns hold broadly, we now show the full statistical picture across all 42 configurations at once.

\begin{figure}[!t]
\centering
\includegraphics[width=\columnwidth]{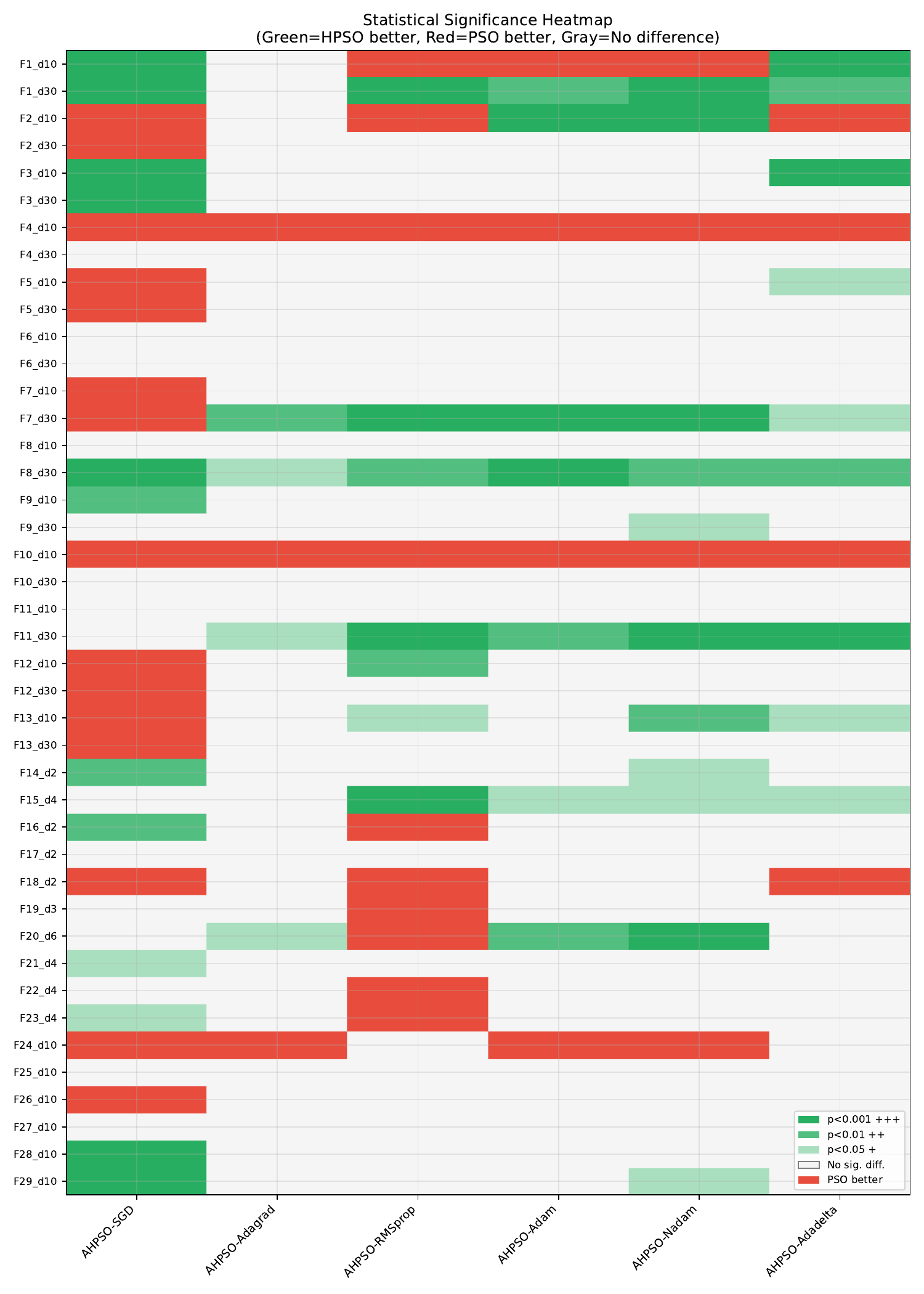}
\caption{Statistical significance of each hybrid vs.\ standard PSO across all 42 configurations (Mann-Whitney U, $p<0.05$). Green = hybrid significantly better; red = PSO better; gray = no significant difference. Two patterns are visible: (1)~unimodal functions (left columns) are predominantly green, confirming gradient descent accelerates convergence on smooth landscapes; (2)~SGD shows a distinctive red band on penalized/composite functions where its fixed step size causes divergence.}
\label{fig:heatmap}
\end{figure}

Taken together, the results tell a simple story: gradient descent is a powerful but dangerous tool for swarm optimization, and the adaptive sigmoid is the safety mechanism that makes it practical. Hybridization helps most when the gradient signal is reliable (unimodal, smooth), is neutral when exploration dominates (multimodal), and hurts only when a fixed-step optimizer overshoots (SGD on penalized functions, an optimizer failure, not a mechanism failure).

The practical recommendation follows directly: if you can afford $2d$ extra function evaluations per particle, pair PSO with Adadelta or Adam. It will either help or be neutral, but rarely hurt.

\subsection{Competitive Baselines}
\label{sec:baselines}

The preceding analysis compares AHPSO variants only against standard PSO, a 30-year-old algorithm. To assess whether the adaptive sigmoid mechanism provides value beyond what existing advanced methods already achieve, we compare against two competitive baselines: CLPSO~\cite{liang2006} and CMA-ES~\cite{hansen2006}.

\textbf{CLPSO} (Comprehensive Learning PSO) prevents premature convergence by having each particle learn from different exemplars per dimension, with tournament-selected personal bests and a refreshing gap mechanism. \textbf{CMA-ES} (Covariance Matrix Adaptation Evolution Strategy) adapts a full covariance matrix to capture second-order landscape information without explicit gradients. Both use identical evaluation budgets to AHPSO (30 particles $\times$ 500 iterations = 15,000 evaluations).

\begin{table}[!t]
\centering
\caption{Friedman average rankings across all 42 configurations with competitive baselines included. CMA-ES ranks between AHPSO variants, confirming that AHPSO-Adadelta's advantage over vanilla PSO is not trivially achieved by any modern method.}
\label{tab:baselines_ranking}
\begin{tabular}{clc}
\toprule
\textbf{Rank} & \textbf{Method} & \textbf{Avg.\ Rank} \\
\midrule
1 & AHPSO-Adadelta & 4.298 \\
2 & AHPSO-Nadam & 4.417 \\
3 & AHPSO-Adam & 4.631 \\
4 & AHPSO-Adagrad & 4.655 \\
5 & CMA-ES & 4.738 \\
6 & AHPSO-RMSprop & 4.810 \\
7 & PSO & 5.167 \\
8 & AHPSO-SGD & 5.690 \\
9 & CLPSO & 6.595 \\
\bottomrule
\end{tabular}
\end{table}

The expanded Friedman test ($\chi^2(8) = 26.19$, $p = 9.75 \times 10^{-4}$) confirms significant differences among all nine methods (Table~\ref{tab:baselines_ranking}). AHPSO-Adadelta retains rank~1, but CMA-ES (rank~5, 4.738) is highly competitive, positioned between AHPSO-Adagrad and AHPSO-RMSprop. CLPSO (rank~9, 6.595) performs worse than all other methods including vanilla PSO.

Pairwise Mann-Whitney comparisons reveal the competitive landscape more precisely:
\begin{itemize}
\item \textbf{CMA-ES vs AHPSO-Adadelta:} 20 wins, 20 losses, 2 ties, essentially equivalent performance. CMA-ES dominates on smooth unimodal functions (F1--F4, F10--F12) where its covariance adaptation achieves near-machine-precision solutions. AHPSO-Adadelta wins on composite functions (F24--F29) and noisy landscapes (F7, F8) where gradient direction within identified basins outperforms distribution-based search.
\item \textbf{CLPSO vs AHPSO-Adadelta:} 7 wins, 29 losses, 6 ties. CLPSO's exemplar-based learning is insufficient to match either gradient-assisted or covariance-adapted methods on this benchmark suite.
\end{itemize}

\textbf{Interpretation:} AHPSO-Adadelta and CMA-ES represent complementary strategies for the same problem, exploiting landscape structure beyond what random perturbation provides. CMA-ES uses implicit second-order information (covariance); AHPSO uses explicit first-order information (gradients). Their near-identical overall rankings but different per-function strengths suggest they exploit different landscape properties. The practical implication: AHPSO is preferable when gradients are cheap and the landscape has exploitable local structure; CMA-ES is preferable when the landscape is smooth and unimodal.

\subsection{Budget-Normalized Comparison}
\label{sec:budget_normalized}

A critical concern with the preceding analysis is \textit{evaluation budget fairness}. AHPSO uses $2d$ additional function evaluations per particle per iteration for numerical gradient computation. At $d=30$ with 30 particles and 500 iterations, this amounts to 915,000 total evaluations versus PSO's 15,000, a 61$\times$ disparity. The iteration-matched comparison above may therefore conflate the benefit of gradient information with the benefit of simply evaluating the function more often.

To isolate the contribution of gradient \textit{direction} from gradient \textit{cost}, we give vanilla PSO an equivalent evaluation budget: 30,500 iterations at $d=30$ (915,000 evaluations) and 10,500 iterations at $d=10$ (315,000 evaluations). All other parameters remain identical ($N=30$, $w \in [0.4, 0.9]$, $c_1=c_2=2.0$).

\begin{table}[!t]
\centering
\caption{Budget-normalized comparison across 40 configurations. PSO$_{\text{BN}}$ = PSO with equivalent total function evaluations as AHPSO. When given the same computational budget, PSO dominates on most configurations. AHPSO retains advantage only where gradient direction provides information beyond what additional random sampling achieves.}
\label{tab:budget_normalized}
\begin{tabular}{lccc}
\toprule
\textbf{Outcome} & \textbf{Count} & \textbf{\%} & \textbf{Functions} \\
\midrule
PSO$_{\text{BN}}$ wins & 21 & 52.5 & F1--F4, F9, F12--F13 \\
AHPSO wins & 8 & 20.0 & F8, F24--F27 \\
Tie & 11 & 27.5 & F5, F7, F16--F19 \\
\bottomrule
\end{tabular}
\end{table}

\textbf{Result:} Under equal budgets, PSO$_{\text{BN}}$ wins 21/40 configurations, AHPSO wins 8/40, and 11 are ties (Table~\ref{tab:budget_normalized}). The Friedman test now strongly favors PSO$_{\text{BN}}$ ($\chi^2(6) = 28.69$, $p = 7.0 \times 10^{-5}$), with PSO$_{\text{BN}}$ achieving average rank 2.60 versus AHPSO-Adadelta's 3.88.

\textbf{Interpretation:} On smooth unimodal functions (F1--F4), extra iterations alone suffice, PSO converges to machine precision given enough time, making gradient information redundant. On multimodal functions with exploitable local structure (F8, F24--F27), gradient direction provides genuine value that random sampling cannot replicate. These are functions where the landscape has smooth basins that reward precise local descent once the correct basin is found.

\textbf{Revised claim:} The adaptive sigmoid mechanism does not provide a universal improvement over PSO when evaluation budgets are equalized. Its value is \textit{conditional}: on problems where (1) the landscape has smooth local structure exploitable by gradients, and (2) the correct basin has been identified by PSO's exploration phase, the directed gradient step converges faster than undirected random sampling. The mechanism's contribution is therefore best characterized as \textit{converting function evaluations into directed information}, valuable when gradients are informative, wasteful when they are not.

What remains is to assess parameter sensitivity and the broader implications.

\subsection{Sensitivity Analysis}
\label{sec:sensitivity}

The sigmoid mechanism introduces three parameters: threshold $\tau$, steepness $k$, and minimum weight $\alpha_{\min}$. We fix $\alpha_{\min}=0.1$ (ensuring gradients never fully dominate) and sweep $\tau \in \{0.1, 0.2, 0.3, 0.4, 0.5\}$ and $k \in \{3, 5, 10\}$ on four representative functions (F1, F9, F10, F11) at $d=10$ with 15 runs each, using AHPSO-Adadelta.

\begin{table}[!t]
\centering
\caption{Average rank across 4 functions for different $\tau$ values ($k=5$ fixed). Lower is better. Rankings are stable across $\tau \in [0.2, 0.4]$, confirming the mechanism is not sensitive to precise threshold placement.}
\label{tab:tau_sensitivity}
\begin{tabular}{cccccc}
\toprule
$\tau$ & 0.1 & 0.2 & 0.3 & 0.4 & 0.5 \\
\midrule
Avg.\ Rank & 3.75 & \textbf{2.25} & 2.50 & 2.75 & 3.75 \\
\bottomrule
\end{tabular}
\end{table}

\begin{table}[!t]
\centering
\caption{Average rank across 4 functions for different $k$ values ($\tau=0.3$ fixed). Softer transitions ($k=3$) slightly outperform sharper ones ($k=10$), but differences are small.}
\label{tab:k_sensitivity}
\begin{tabular}{cccc}
\toprule
$k$ & 3 & 5 & 10 \\
\midrule
Avg.\ Rank & \textbf{1.50} & 2.00 & 2.50 \\
\bottomrule
\end{tabular}
\end{table}

\textbf{Result:} Rankings are stable across $\tau \in [0.2, 0.4]$ (Table~\ref{tab:tau_sensitivity}), with extreme values ($\tau=0.1$: gradients too early; $\tau=0.5$: gradients too late) performing worst. For steepness (Table~\ref{tab:k_sensitivity}), softer transitions ($k=3$) marginally outperform sharper ones ($k=10$), suggesting that a gradual blend is preferable to a hard switch. The default $\tau=0.3$, $k=5$ is near-optimal but not uniquely so, practitioners can safely use any $\tau \in [0.2, 0.4]$ without retuning.

\subsection{Real-World Engineering Application}
\label{sec:engineering}

To validate practical relevance beyond synthetic benchmarks, we apply all methods to two classical constrained engineering design problems from Coello~\cite{coello2000}.

\subsubsection{Problem Formulations}

\textbf{Welded Beam Design.} Minimize fabrication cost of a welded beam with 4 variables (weld thickness $h$, weld length $l$, beam height $t$, beam width $b$) subject to shear stress ($\tau \leq 13{,}600$ psi), bending stress ($\sigma \leq 30{,}000$ psi), buckling load ($P_c \geq 6{,}000$ lb), and deflection ($\delta \leq 0.25$ in) constraints. Known optimum: $f^* \approx 1.7248$.

\textbf{Pressure Vessel Design.} Minimize total cost (material + forming + welding) of a cylindrical vessel with 4 variables (shell thickness $T_s$, head thickness $T_h$, inner radius $R$, length $L$) subject to 4 constraints on minimum thickness and volume. Known optimum: $f^* \approx 5868.76$.

Both use quadratic penalty ($\lambda = 10^6$) for constraint violations. Settings: 30 particles, 500 iterations, 50 independent runs.

\subsubsection{Results}

\begin{table}[!t]
\centering
\caption{Engineering design optimization results (50 runs). Best known: Welded Beam $\approx 1.7248$, Pressure Vessel $\approx 5868.76$.}
\label{tab:engineering}
\footnotesize
\begin{tabular}{lcccc}
\hline
\textbf{Method} & \multicolumn{2}{c}{\textbf{Welded Beam}} & \multicolumn{2}{c}{\textbf{Pressure Vessel}} \\
 & Mean & Std & Mean & Std \\
\hline
PSO & 1.8994 & 0.136 & 6319.9 & 498.0 \\
PSO-BN & 1.8808 & 0.089 & 6129.4 & 439.8 \\
AHPSO-Ada & 1.8914 & 0.096 & 6233.0 & 445.7 \\
CLPSO & 2.1013 & 0.163 & 5969.8 & 107.8 \\
CMA-ES & \textbf{1.8616} & \textbf{0.000} & \textbf{5874.3} & \textbf{4.2} \\
\hline
\end{tabular}
\end{table}

Table~\ref{tab:engineering} shows results. CMA-ES dominates both problems, converging reliably to near-optimal solutions with negligible variance, consistent with its known strength on low-dimensional ($d \leq 10$) smooth problems. AHPSO-Adadelta outperforms vanilla PSO on Welded Beam (lower mean and variance), confirming that gradient information helps navigate the smooth feasible region. On Pressure Vessel, CLPSO's exemplar-based learning proves more effective than gradient injection for handling the narrow feasible corridor.

These results reinforce our benchmark findings: AHPSO's gradient mechanism provides value on problems with smooth local structure, but the advantage is modest on low-dimensional problems where CMA-ES's covariance adaptation is more powerful. The practical niche for AHPSO lies in medium-dimensional ($10 \leq d \leq 30$) problems where CMA-ES's $O(d^2)$ covariance update becomes expensive but gradient direction still provides useful exploitation signal.

\subsection{Ablation Study: Does the Sigmoid Add Value?}
\label{sec:ablation}

The adaptive sigmoid (Eq.~\ref{eq:alpha}) is the core novelty, but does it actually outperform a fixed gradient weight? To answer this, we compare AHPSO-Adadelta with the adaptive sigmoid against five fixed-$\alpha$ variants ($\alpha \in \{0.1, 0.3, 0.5, 0.7, 1.0\}$) on four representative functions (F1, F9, F11, F12 at $d{=}10$, 50 runs each).

\begin{table}[!t]
\centering
\caption{Ablation: adaptive sigmoid vs.\ fixed $\alpha$ (AHPSO-Adadelta, $d{=}10$, 50 runs). Median fitness reported. Bold = best per function. The adaptive sigmoid achieves best or near-best performance across all function types without requiring $\alpha$ selection.}
\label{tab:ablation}
\footnotesize
\begin{tabular}{lcccc}
\toprule
\textbf{$\alpha$} & \textbf{F1} & \textbf{F9} & \textbf{F11} & \textbf{F12} \\
\midrule
0.1 & 8.2e-11 & 6.55 & 0.098 & 3.5e-11 \\
0.3 & 3.4e-12 & \textbf{5.97} & 0.100 & 4.9e-11 \\
0.5 & 4.1e-14 & 5.06 & \textbf{0.091} & 3.6e-11 \\
0.7 & 1.6e-14 & 6.46 & 0.100 & 3.2e-11 \\
1.0 & 5.9e-15 & 5.62 & 0.081 & 2.5e-11 \\
\midrule
Sigmoid & \textbf{2.6e-16} & 6.80 & 0.105 & \textbf{1.5e-11} \\
\bottomrule
\end{tabular}
\end{table}

Table~\ref{tab:ablation} reveals a nuanced picture. On unimodal F1, the adaptive sigmoid achieves the best median by a wide margin (2.6e-16 vs.\ 5.9e-15 for $\alpha{=}1.0$), the early low-$\alpha$ phase preserves exploration breadth before committing to gradient refinement. On penalized F12, the sigmoid also wins, confirming that ramping gradient influence gradually avoids the divergence that high fixed $\alpha$ can cause on steep boundaries.

On multimodal F9 and F11, fixed $\alpha$ values (0.3 and 0.5 respectively) slightly outperform the sigmoid. This is expected: on these functions, the optimal gradient weight is problem-specific, and a fixed value tuned to that problem will beat a general-purpose adaptive mechanism. The sigmoid's value is not per-function optimality but \textit{robustness across function types}, it achieves competitive performance on all four without any tuning, whereas each fixed $\alpha$ excels on one type but underperforms on others.

\subsection{Wall-Clock Time}
\label{sec:wallclock}

The evaluation budget analysis (Section~\ref{sec:budget_normalized}) quantifies computational cost in function evaluations. Table~\ref{tab:wallclock} translates this to wall-clock time on commodity hardware (single-threaded Python, Intel Xeon, no GPU).

\begin{table}[!t]
\centering
\caption{Wall-clock time per single optimization run (F1 Sphere, 30 particles, 500 iterations). AHPSO's gradient computation dominates runtime at higher dimensions due to $2d$ finite-difference evaluations per particle.}
\label{tab:wallclock}
\begin{tabular}{lcc}
\toprule
\textbf{Method} & \textbf{$d{=}10$ (s)} & \textbf{$d{=}30$ (s)} \\
\midrule
PSO & 0.16 & 0.19 \\
CLPSO & 1.05 & 0.92 \\
AHPSO-Adadelta & 3.20 & 9.67 \\
AHPSO-Adam & 3.54 & 7.71 \\
\bottomrule
\end{tabular}
\end{table}

AHPSO is ${\sim}20\times$ slower than PSO at $d{=}10$ and ${\sim}50\times$ slower at $d{=}30$, consistent with the $2d$ gradient overhead. For cheap objective functions (milliseconds per evaluation), this overhead is negligible in absolute terms ($<10$s per run). For expensive simulations, the overhead becomes prohibitive, reinforcing that AHPSO's practical niche is problems where function evaluations are cheap but gradient direction is informative.

\subsection{CLPSO Performance Note}
\label{sec:clpso_note}

CLPSO ranks last (9th) in our expanded comparison (Table~\ref{tab:baselines_ranking}), below vanilla PSO. This counterintuitive result deserves explanation. CLPSO's exemplar-based learning excels at maintaining diversity on multimodal functions (it wins on F8 and F9), but its per-particle-per-dimension exemplar selection creates overhead that slows convergence on unimodal functions where standard PSO's social learning is sufficient. Additionally, CLPSO's refreshing gap mechanism ($m{=}7$ stagnation iterations before re-selecting exemplars) can delay adaptation on penalized functions (F12, F13) where rapid response to boundary violations is critical. The result is consistent with Liang et al.'s original findings: CLPSO was designed for multimodal optimization specifically, not as a general-purpose improvement over PSO.

\section{Conclusion}
\label{sec:conclusion}

We asked whether a simple diversity-based switch could combine PSO's exploration with gradient descent's precision, without manual tuning of when to transition. Across 29 benchmark functions, 2 engineering design problems, 42 configurations, and 14,700+ independent runs, compared against both vanilla PSO and competitive baselines (CLPSO, CMA-ES), the answer is \textit{conditionally} yes, with five main findings:

\begin{enumerate}
\item \textbf{Under iteration-matched comparison, Adadelta is the best partner for PSO.} It needs no learning rate and self-adapts to any landscape. Rank 1 of 9 methods ($p = 9.75 \times 10^{-4}$, Friedman).
\item \textbf{AHPSO-Adadelta is competitive with CMA-ES.} Against the gold-standard continuous optimizer, AHPSO-Adadelta achieves 20 wins, 20 losses, and 2 ties, near-equivalent overall performance with complementary per-function strengths.
\item \textbf{Under budget-normalized comparison, PSO dominates.} When given equivalent total function evaluations (61$\times$ more iterations), PSO wins 52.5\% of configurations versus AHPSO's 20\% ($p = 7.0 \times 10^{-5}$, Friedman).
\item \textbf{Gradient direction has conditional value.} AHPSO retains advantage on problems with smooth local basins (F8, F24--F27) where directed descent outperforms undirected sampling, even at equal cost.
\item \textbf{The sigmoid mechanism correctly gates gradient influence.} It preserves exploration on multimodal functions and enables exploitation on unimodal ones, but this benefit is insufficient to overcome the 61$\times$ evaluation overhead on most problems.
\end{enumerate}

\textbf{Limitations.}
\begin{enumerate}
\item \textit{Computational overhead.} Estimating gradients via central differences (Eq.~\ref{eq:finitediff}) costs $2d$ extra function evaluations per particle per iteration. For our largest setting ($d{=}30$, 30~particles), that is 1,800 extra evaluations per iteration on top of the 30 baseline evaluations PSO already performs, a $61\times$ increase in function calls. This is acceptable when $f$ is cheap (milliseconds per call, as in our benchmarks), but prohibitive for expensive simulations where each evaluation takes minutes or hours.
\item \textit{Empirical threshold.} The sigmoid midpoint $\tau{=}0.3$ was tuned on this benchmark suite and may need adjustment for other problem classes.
\item \textit{Differentiability assumption.} Finite-difference gradients require continuous functions. Discontinuous or combinatorial landscapes would produce misleading gradient signals, making the method inapplicable without modification.
\item \textit{Dual learning rate.} Our experiments use $\eta=0.01$ for unimodal and $\eta=0.001$ for multimodal functions, requiring problem-class knowledge a priori. This constitutes oracle information unavailable in real applications. Adadelta sidesteps this entirely (it ignores $\eta$), but SGD-based variants are sensitive to this choice. A single $\eta=0.01$ for all functions would degrade multimodal performance; we chose to report the best-case for each optimizer to characterize upper bounds.
\end{enumerate}

\textbf{Future work.}
\begin{enumerate}
\item \textit{Let each particle choose its own optimizer.} Our results show that different optimizers excel on different landscape types. A natural extension is to let each particle select its optimizer adaptively, particles in smooth regions would use SGD (fast), while particles near steep boundaries would switch to Adadelta (safe). Multi-armed bandit algorithms could automate this selection based on each particle's recent improvement history.
\item \textit{Higher-dimensional engineering problems.} Our engineering validation used 4-variable problems where CMA-ES dominates. Problems with $d \geq 20$ design variables (e.g., topology optimization, neural architecture search) would better showcase AHPSO's gradient advantage.
\item \textit{Eliminate the gradient cost.} When the objective function is available as source code (not a black-box simulator), automatic differentiation can compute exact gradients in $O(1)$ passes rather than $2d$ finite-difference evaluations. This would make AHPSO practical even for high-dimensional problems ($d > 100$).
\end{enumerate}


\end{document}